\documentclass[10pt,twocolumn,letterpaper]{article}
\usepackage[pagenumbers]{cvpr}
\usepackage{microtype}
\usepackage{array}
\usepackage{colortbl}
\usepackage{tabularx}
\newcommand{\method}{AdaREPA\xspace}
\newcommand{\fm}{\mathcal{L}_{\mathrm{FM}}}
\newcommand{\aux}{\mathcal{L}_{\mathrm{REPA}}}
\newcommand{\norm}[1]{\lVert#1\rVert}

\usepackage{float}
\newfloat{algorithm}{tbp}{loa}
\floatname{algorithm}{Algorithm}

\providecommand{\method}{AdaREPA\xspace}
\renewcommand{\method}{AdaREPA\xspace}

\graphicspath{{fig/}}
\definecolor{cvprblue}{rgb}{0.21,0.49,0.74}
\usepackage[breaklinks,colorlinks,allcolors=cvprblue]{hyperref}
\def\confName{CVPR}
\def\confYear{2027}
\title{Where Should a Diffusion Transformer Meet Its Teacher?}
\author{%
Zhuoqi Hu\textsuperscript{1,}\thanks{Both authors contributed equally to this work.}\quad
Shiwei Li\textsuperscript{2,3,*}\quad
Haozhao Wang\textsuperscript{1}\quad
Xiandi Luo\textsuperscript{1}\quad
Yihao Ouyang\textsuperscript{1}\\[3pt]
Wenting Shen\textsuperscript{2}\quad
Ang Wang\textsuperscript{2}\quad
Jiamang Wang\textsuperscript{2}\quad
Minyi Guo\textsuperscript{3}\quad
Ruixuan Li\textsuperscript{1}\\[6pt]
{\normalsize \textsuperscript{1}Huazhong University of Science and Technology}\\
{\normalsize \textsuperscript{2}Alibaba Group\qquad
\textsuperscript{3}Shanghai Jiao Tong University}%
}

\hypersetup{pdfauthor={Zhuoqi Hu, Shiwei Li, Haozhao Wang, Xiandi Luo, Yihao Ouyang, Wenting Shen, Ang Wang, Jiamang Wang, Minyi Guo, Ruixuan Li},pdftitle={Where Should a Diffusion Transformer Meet Its Teacher?}}
\begin{document}
\maketitle

\begin{abstract}
Representation Alignment (REPA) significantly accelerates Diffusion Transformer training by guiding intermediate student features with a pretrained visual teacher, yet existing methods typically attach supervision to a single fixed layer. As student representations evolve, however, we find that the most effective layer for alignment varies across training stages. To track this evolving layer suitability online without costly search, we measure gradient compatibility between the flow-matching and alignment objectives, and propose \textbf{\method}, an online routing framework that adaptively selects the supervision layer as training progresses. AdaREPA uses a softmax distribution over smoothed layer compatibility scores to sample a layer for alignment at each training step. This probabilistic routing adapts teacher supervision to the student's changing learning needs, with minimal training overhead and no additional inference cost. On ImageNet-256, \method accelerates convergence and improves generation quality across model sizes, while revealing how the most suitable layers for supervision shift over training.
\end{abstract}

\section{Introduction}
\label{sec:intro}
\begin{figure}[t]
\centering\includegraphics[width=\linewidth]{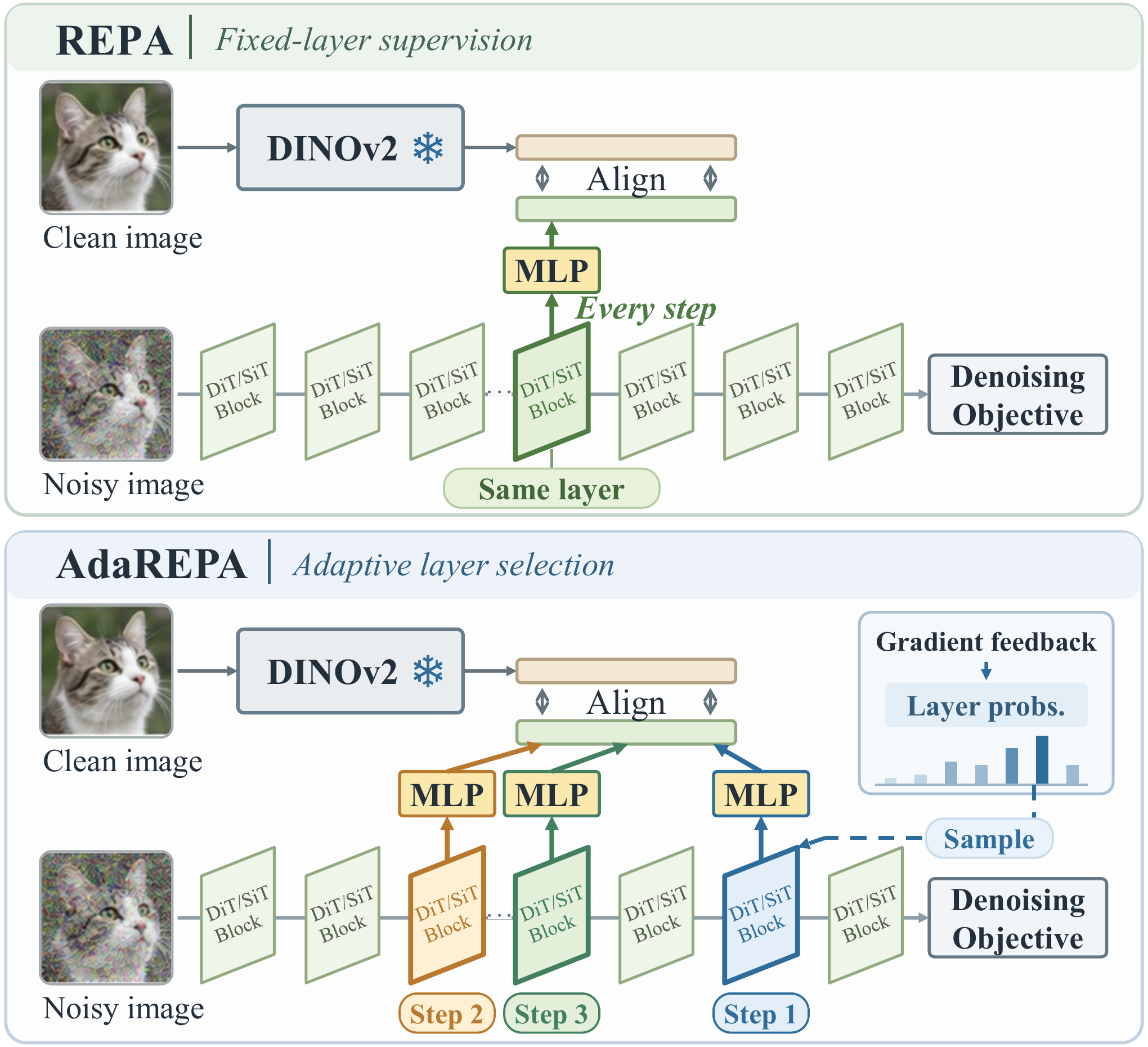}
\caption{\textbf{From fixed-layer supervision to adaptive layer selection.} REPA (top) supervises the same student layer throughout training. \method (bottom) uses gradient compatibility to update layer-selection probabilities and samples a layer for supervision at each training step.}
\label{fig:teaser}
\end{figure}

Diffusion Transformers (such as DiT~\cite{dit} and SiT~\cite{sit}) have achieved remarkable success in visual generation, yet training them demands immense computational resources. To improve training efficiency, Representation Alignment (REPA)~\cite{repa} aligns the model's intermediate features with clean representations from a frozen, self-supervised visual encoder~\cite{dino}, providing mature semantic guidance rather than learning representations from scratch. Subsequent studies~\cite{irepa,rare,haste} have largely followed this paradigm by manually selecting a fixed intermediate layer (e.g., Layer 4 or Layer 8) and keeping this alignment depth fixed throughout training. Intuitively, however, as training progresses, representations across layers evolve dynamically. To investigate whether this static alignment is truly optimal, we systematically evaluate different fixed alignment depths throughout training and compare their effects when introduced from shared SiT checkpoints at different stages (Sec.~\ref{sec:motivation}). Our experiments show that different layers perform better at different stages, and the preferred supervision layer changes with the model state. These findings indicate that confining alignment to a single static layer bottlenecks training potential, raising a fundamental question: \emph{as generative representations evolve over time, which layer should receive teacher guidance at each training stage?}

Addressing this question via offline grid search or manual layer selection is computationally prohibitive for large-scale generative models, calling instead for a lightweight online signal to evaluate layer suitability during training. Inspired by classic principles in multi-task learning~\cite{du,pcgrad}, gradient alignment provides a first-order measure of how an auxiliary update affects the primary objective. We therefore track the interaction between Flow Matching (FM)~\cite{fm,rectified,sit} and REPA at intermediate hidden states, using our Cauchy--Schwarz Cosine (CS-Cosine) score to compare layer compatibility (Sec.~\ref{sec:feedback}). Building on these scores, we propose \textbf{\method} (\textbf{Ada}ptive-\textbf{REPA}), an online routing framework that adapts the supervision layer throughout training (Fig.~\ref{fig:teaser}). A Softmax distribution turns relative layer scores into sampling probabilities, favoring compatible depths while retaining opportunities to revisit other layers. At each step, we sample one layer from this distribution and align its representation with the teacher (Sec.~\ref{sec:selection}). The probabilities thus control how often each layer receives guidance, allowing supervision to follow the student's changing learning needs.

Experiments on class-conditional ImageNet-256~\cite{imagenet} across four SiT model sizes show consistent improvements over REPA. At 400K steps, \method reduces unguided FID from 21.960 to 19.848 on SiT-B/2. With classifier-free guidance at scale 1.5, it achieves the lowest FID among the compared methods across all four sizes, reaching 2.728 on SiT-XL/2 (Table~\ref{tab:benchmark}). Notably, without a prescribed schedule, the learned routing shifts supervision across layers as training progresses (Sec.~\ref{sec:routing-dynamics}), consistent with the stage-dependent layer preferences in Sec.~\ref{sec:motivation}.

Our contributions are:
\begin{itemize}[leftmargin=*,itemsep=1pt,topsep=2pt]
\item We show that the most effective REPA supervision layer changes with the student's training state, motivating adaptive layer selection throughout training.
\item We introduce \method, an online layer selection method that uses FM--REPA gradient compatibility to adjust how often each student layer receives teacher supervision.
\item Experiments across model sizes and visual teachers show improved generation quality over fixed-layer REPA. We further analyze the learned supervision probabilities and the effects of gradient scoring and projector design.
\end{itemize}

\section{Related Work}
\label{sec:related}
\paragraph{Representation alignment in visual generation.}
Building on REPA~\cite{repa}, recent studies explore \emph{what} features to align and \emph{when} to intervene: iREPA~\cite{irepa} designs spatial normalization and convolutional heads to preserve 2D topology; RARE~\cite{rare} selects a teacher layer based on recoverability gaps and adjusts token weights and alignment duration using training-time residuals; and HASTE~\cite{haste} introduces stage-wise early stopping. In these designs, \emph{where} to inject supervision across the backbone remains manually specified. Complementing these choices of teacher targets and supervision duration, \method adapts the student layer that receives teacher guidance as training progresses.

\paragraph{Intermediate feature supervision and layer selection.}
Intermediate supervision has long been used to improve representation learning, from Deeply-Supervised Nets~\cite{dsn} to feature and attention transfer~\cite{fitnets,attentiontransfer}. Cross-layer distillation methods, such as SemCKD~\cite{semckd} and ALP-KD~\cite{alpkd}, learn weighted correspondences between student and teacher features. ReviewKD~\cite{reviewkd} uses cross-stage feature fusion to draw supervision from multiple teacher depths. In audio flow matching, AG-REPA~\cite{agrepa} selects alignment layers using gate-ablation sensitivity during an offline warm-up phase and keeps this selection fixed for subsequent training. \method instead revisits the supervision position throughout training through an evolving sampling distribution, while aligning one student layer per step.

\paragraph{Gradient similarity in auxiliary learning.}
Gradient information guides loss weighting and update directions in auxiliary and multi-task learning. Du~\etal~\cite{du} weight auxiliary updates using their cosine similarity with the primary gradient, while GradNorm~\cite{gradnorm} balances task contributions by adjusting gradient magnitudes. To address conflicting objectives, MGDA~\cite{mgda} seeks a common descent direction, PCGrad~\cite{pcgrad} projects away conflicting gradient components, and CAGrad~\cite{cagrad} regularizes average-loss optimization using the worst local task improvement. Gradient Vaccine~\cite{gradvaccine} encourages aligned updates between related tasks. In diffusion alignment, HASTE~\cite{haste} analyzes how alignment and denoising gradients become less compatible over training, motivating stage-wise termination of teacher supervision. These studies motivate using gradient agreement to guide auxiliary supervision, and \method uses this signal to adjust where teacher guidance enters the student over training.

\section{Method}
\label{sec:method}
We first establish the motivation for adaptive layer selection through three experiments, revealing the limitations of static alignment and the relationship between gradient similarity and layer suitability (Sec.~\ref{sec:motivation}). We then define a gradient compatibility score for comparing student layers (Sec.~\ref{sec:feedback}) and use it to guide adaptive layer selection (Sec.~\ref{sec:selection}).

\subsection{Motivation}
\label{sec:motivation}
The alignment layer determines which representation receives teacher constraints and how far alignment gradients propagate through the backbone. As student representations evolve, a fixed position may become less suitable for the current learning needs. We examine fixed-layer trajectories, supervision introduced from shared model states, and the corresponding gradient relationships to motivate adaptive layer selection.

\subsubsection{Stage-Dependent Effects of Fixed Supervision}
\label{sec:reversal}
We compare SiT-S/2 generation trajectories with REPA applied separately at Layers 2, 4, and 8 from the start, covering different supervision depths. The choice of supervision layer substantially affects generation quality. Layer 4 achieves the best final FID, while Layer 8 gives slightly lower FID at 50K, 200K, and 300K (Table~\ref{tab:fixed-layer-trajectories}). Since the supervision position remains fixed throughout training, these results reflect how that position shapes the full learning trajectory. We therefore examine the role of supervision depth in representation learning and generative refinement. REPA~\cite{repa} suggests that alignment at earlier layers lets later blocks learn high-frequency details from improved representations. However, in our comparison, Layer 2 leaves more blocks after alignment but performs worse than Layer 4 at most listed checkpoints. HASTE~\cite{haste} further shows that stopping alignment after a suitable training stage can improve subsequent generation. Together, these findings motivate considering supervision depth in relation to the student's training stage.

\begin{table}[t]
\centering
\caption{\textbf{FID-10K ($\downarrow$) with fixed supervision layers.}}
\label{tab:fixed-layer-trajectories}
\small
\setlength{\tabcolsep}{2pt}
\begin{tabular*}{\columnwidth}{@{\extracolsep{\fill}}lrrrrrr@{}}
\toprule
Layer & 50K & 100K & 150K & 200K & 300K & 400K \\
\midrule
2 & 108.880 & 90.974 & 81.324 & 71.563 & 66.968 & 60.224 \\
4 & 106.043 & \textbf{87.029} & \textbf{79.177} & 71.764 & 58.792 & \textbf{51.795} \\
8 & \textbf{105.762} & 93.165 & 79.779 & \textbf{71.049} & \textbf{58.572} & 58.137 \\
\bottomrule
\end{tabular*}
\end{table}

\begin{figure}[t]
\centering
\includegraphics[width=\linewidth]{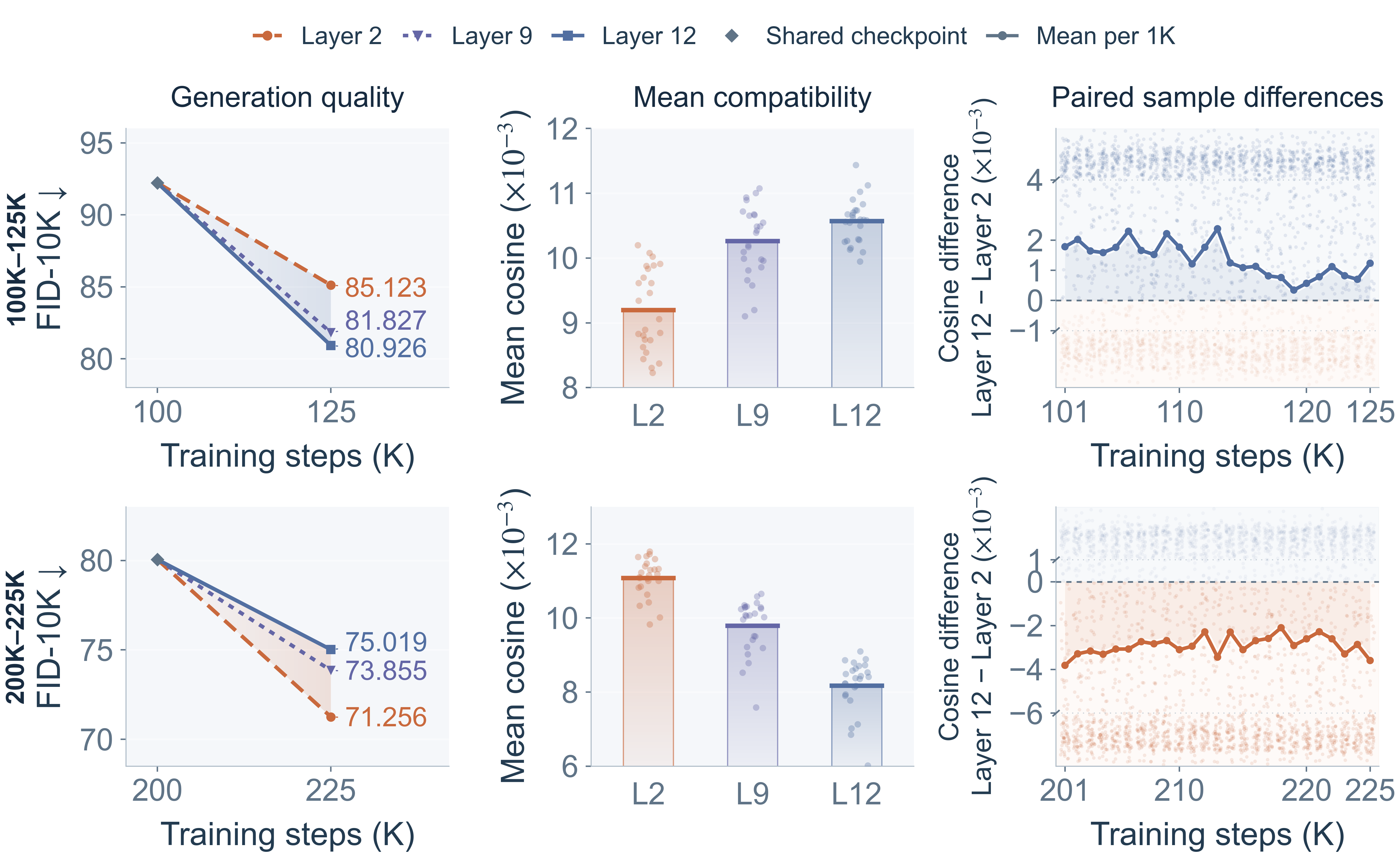}
\caption{\textbf{Stage-dependent responses to REPA supervision.} Rows show 25K continuations from shared SiT-S/2 checkpoints at 100K and 200K. Left: FID-10K at Layers 2, 9, and 12. Middle: Mean gradient cosine similarity. Right: Paired per-sample cosine differences (Layer 12 minus Layer 2).}
\label{fig:stage-response}
\end{figure}

\begin{figure*}[t]
    \centering
    \includegraphics[width=0.95\textwidth]{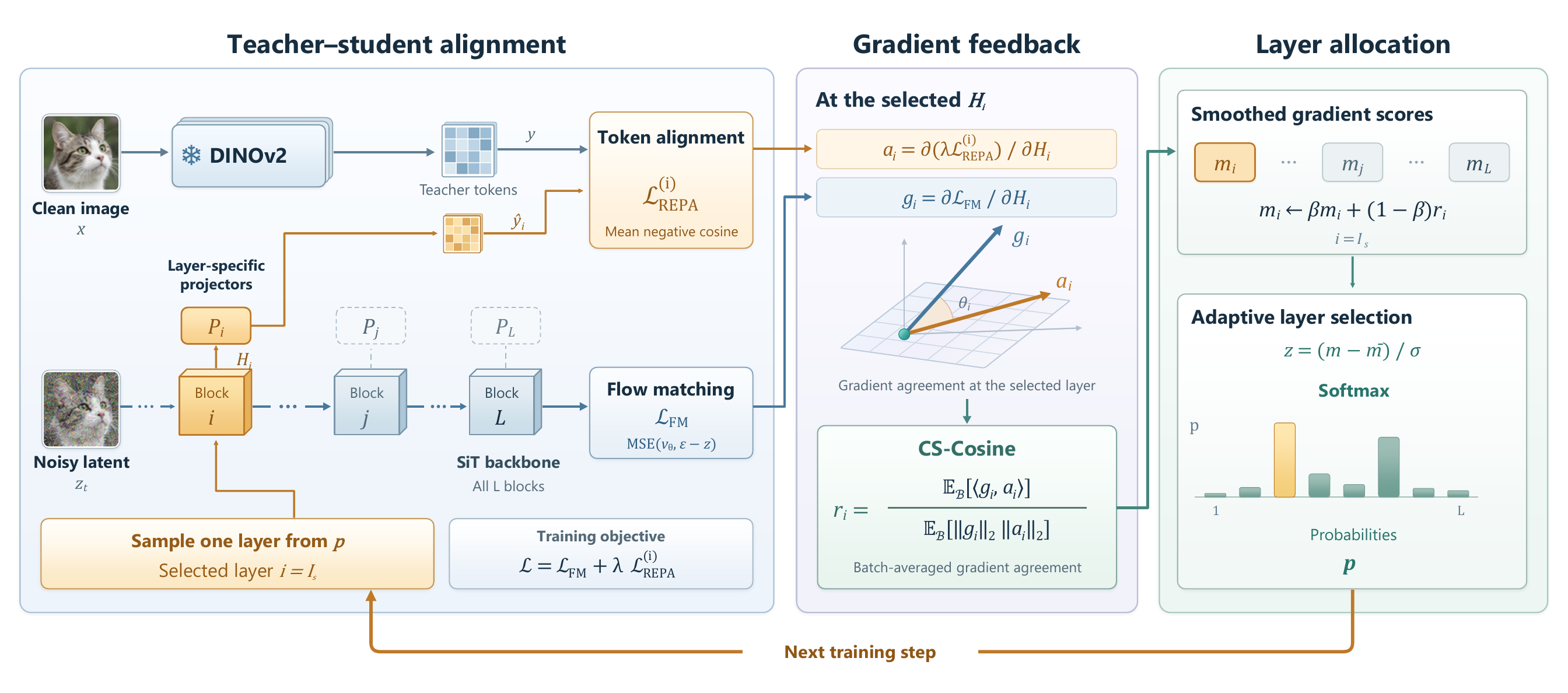}
    \caption{
    \textbf{Overview of \method.}
    One sampled student layer is aligned with the frozen teacher through its own projector.
    FM and REPA gradients at the selected hidden state provide a CS-Cosine score.
    Smoothed scores are standardized and mapped to Softmax probabilities for subsequent layer selection.
    }
    \label{fig:method}
\end{figure*}

We hypothesize that the benefit of a supervision layer depends on both the student's current semantic representations and the capacity left in subsequent blocks for generative refinement. Shallower supervision leaves more blocks for generative refinement, but this advantage relies on useful representations at the supervision layer. Deeper supervision may help more preceding blocks learn semantic structure from the teacher, but leaves fewer subsequent blocks for generative refinement. As student representations evolve, the balance between these roles may change. A relatively intermediate layer may therefore offer a good overall compromise and yield better final performance under fixed supervision. However, the changing balance between semantic learning and generative refinement raises a question about adaptive layer selection: which layer is most suitable for supervision at each individual training stage?

\subsubsection{Introducing Supervision at Different Model States}
\label{sec:stage-preferences}
To examine this question, we compare supervision depths from shared checkpoints at different stages. We first train SiT-S/2 using only the flow matching (FM) objective. During this training, a separate projector at each layer learns to align detached SiT features with teacher representations. Alignment gradients update the projectors without propagating into SiT. At 100K and 200K, the student has accumulated different learning histories. We use the shallow Layer 2 and the final Layer 12 as the main comparison. From each full checkpoint, we introduce REPA with $\lambda=0.5$ at either layer and train for another 25K steps with other settings fixed.

Starting from 100K, the deeper Layer 12 achieves FID-10K of 80.926, compared with 85.123 for Layer 2. Starting from 200K, the ordering reverses: Layer 2 reaches 71.256, compared with 75.019 for Layer 12 (Fig.~\ref{fig:stage-response}, left). This reversal leads us to add Layer 9, a deeper position before the final layer, to examine whether the stage-dependent difference is specific to the network endpoint. Layer 9 also outperforms Layer 2 from 100K (81.827 vs.\ 85.123) but falls behind it from 200K (73.855 vs.\ 71.256). This change therefore extends to a non-final deep layer. These results show that the most suitable supervision layer changes with the student's training state. We therefore wonder what makes a layer more suitable for supervision at a particular training stage.

\subsubsection{Gradient Compatibility and Layer Suitability}
\label{sec:probe}
FM trains the velocity field needed for generation, while REPA provides auxiliary representation guidance. Auxiliary learning studies relate gradient agreement to the change in the main-task loss under a local auxiliary update~\cite{du}. We therefore ask whether this agreement reflects the benefits of supervision at different layers.

In the branches above, we measure the pure FM and active-layer REPA gradients at the layer output every 1K steps, using the same 32 images with four fixed diffusion timesteps per image. We average their per-sample cosine similarities over these 128 observations at each measurement, then average the 25 measurements within each window. In Fig.~\ref{fig:stage-response} (middle), Layer 12 has higher mean cosine similarity in 100K--125K, whereas Layer 2 is higher in 200K--225K. Layer 9 follows the same ordering relative to Layer 2: it has higher mean cosine similarity in the earlier window and lower similarity in the later one. Across all three layers, the ordering of mean gradient agreement matches that of generation quality in both windows.

The right column pairs per-sample cosines from the two branches at the same training step, image, and diffusion timestep, plotting Layer 12 minus Layer 2; pale points show individual differences, and the connected solid points show their mean at each measurement. These means remain positive throughout 101K--125K and negative throughout 201K--225K, despite variation across individual observations. Thus, the stage-dependent preference persists during continued training, supporting gradient agreement as a signal for adapting supervision.

\subsection{Gradient Compatibility Scoring}
\label{sec:feedback}
Based on the findings above, we use gradient similarity to assess each student layer's suitability for alignment. The batch score should retain samples' relative contributions to the local FM loss change.

Let $L$ be the number of student layers and $H_i$ the hidden representation at student layer $i$. The gradients of the flow matching (FM) and REPA losses at $H_i$ are $g_i=\partial\fm/\partial H_i$ and $a_i=\partial(\lambda\aux^{(i)})/\partial H_i$. For a small REPA update $H_i-\eta a_i$ with $\eta>0$, a first-order expansion of the FM loss gives
\begin{equation}
 \fm(H_i-\eta a_i)-\fm(H_i)
 =-\eta\langle g_i,a_i\rangle+\mathcal{O}(\eta^2).
 \label{eq:local}
\end{equation}
The inner product gives each sample's contribution to the first-order FM loss decrease per unit step.

Mean Cosine averages per-sample cosines equally, ignoring differences in gradient magnitudes and thus the samples' relative contributions to the local FM loss change. Batch-Pooled Cosine (BP-Cosine) computes cosine similarity after concatenating gradients across the batch. When FM gradients are larger on one sample and REPA gradients on another, its score can be low even if every sample's gradient pair is aligned. The score thus reflects norm mismatch as well as directional agreement. To retain relative sample contributions without this extra scaling, we average the sample inner products and normalize this average by its maximum at the observed per-sample gradient norms. This maximum is reached when every gradient pair is aligned. We define Cauchy--Schwarz Cosine (CS-Cosine) as
\begin{equation}
 r_i=\frac{\mathbb{E}_{\mathcal B}[\langle g_i,a_i\rangle]}
 {\mathbb{E}_{\mathcal B}[\norm{g_i}_2\norm{a_i}_2]}.
 \label{eq:cs-cosine}
\end{equation}
Here $\mathbb{E}_{\mathcal B}$ denotes the batch average. The Cauchy--Schwarz inequality gives the denominator as this maximum. For a positive denominator, $r_i$ lies in $[-1,1]$ and is unchanged by a common positive rescaling of either gradient.

CS-Cosine weights each sample's cosine by its gradient norm product. This preserves relative sample contributions while removing the extra scaling caused by norm mismatch, providing a score for assessing layer suitability.

\subsection{Adaptive Layer Selection}
\label{sec:selection}
To guide layer selection over training, we smooth the observed compatibility scores with an exponential moving average. This reduces fluctuations between updates while tracking changes in layer suitability:
\begin{equation}
 m_i^{(s+1)}=\beta m_i^{(s)}+(1-\beta)r_i^{(s)},
 \label{eq:ema}
\end{equation}
where $s$ indexes training steps. Each layer's first valid observation initializes its smoothed gradient score, which is updated with new observations when the layer is selected again. Once every layer has a score, we apply Softmax to the standardized scores to obtain sampling probabilities:
\begin{equation}
 \begin{aligned}
 z_i^{(s)}&=\frac{m_i^{(s)}-\bar m^{(s)}}{\sigma^{(s)}},\\
 p_i^{(s)}&=\frac{\exp(z_i^{(s)})}{\sum_{j=1}^{L}\exp(z_j^{(s)})},
 \end{aligned}
 \label{eq:policy}
\end{equation}
where $\bar m^{(s)}$ and $\sigma^{(s)}$ are the mean and standard deviation of the smoothed scores. This yields a sampling distribution that reflects the changing suitability of student layers for teacher supervision as training progresses.

At step $s$, we sample one layer $I_s\sim p^{(s)}$, shared across the batch. Each student layer has its own projector, and the selected projector maps $H_{I_s}$ to the frozen teacher's representation of the clean image. We combine its alignment loss with the FM loss:
\begin{equation}
 \mathcal{L}=\fm+\lambda\aux^{(I_s)},
 \label{eq:loss}
\end{equation}
where $\aux^{(I_s)}$ is the standard REPA cosine loss at the sampled layer and both losses are averaged over the batch.

Overall, \method samples a supervision layer according to the current probabilities, aligns its representation with the teacher, and optimizes the joint FM and REPA objective. Gradient agreement at the selected layer updates its smoothed score and the probabilities for subsequent steps. This feedback loop adjusts how often each depth receives teacher guidance as the student learns.

\section{Experiments}
\label{sec:exp}
We evaluate \method on class-conditional ImageNet-256. We examine generation quality, convergence, and the learned routing trajectory, then ablate batch-wise versus sample-wise selection and design choices. Finally, we compare generated samples over training to examine changes in structure and local detail.

\begin{table*}[t]\centering
\caption{\textbf{ImageNet-256 generation across SiT scales at 400K training steps.} We report FID, Inception Score (IS), Precision (Prec.), and Recall (Rec.) on 50K samples using EMA weights at CFG scales 1 and 1.5. Best and second-best results within each model scale and CFG setting are bold and underlined, respectively.}
\label{tab:benchmark}
\begingroup\small
\definecolor{benchmarkBlue}{HTML}{EAF1F8}
\definecolor{benchmarkGreen}{HTML}{F3F8F5}
\definecolor{benchmarkOrange}{HTML}{FCF0E6}
\setlength{\tabcolsep}{1.2pt}\renewcommand{\arraystretch}{1.20}
\begin{tabularx}{\textwidth}{l*{4}{>{\hsize=1.01\hsize\linewidth=\hsize\raggedleft\arraybackslash}X>{\hsize=1.19\hsize\linewidth=\hsize\raggedleft\arraybackslash}X>{\hsize=0.94\hsize\linewidth=\hsize\raggedleft\arraybackslash}X>{\hsize=0.86\hsize\linewidth=\hsize\raggedleft\arraybackslash}X}}\toprule
\textbf{CFG = 1} & \multicolumn{4}{c}{\textbf{SiT-S/2}} & \multicolumn{4}{c}{\textbf{SiT-B/2}} & \multicolumn{4}{c}{\textbf{SiT-L/2}} & \multicolumn{4}{c}{\textbf{SiT-XL/2}}\\
\cmidrule(lr){2-5}\cmidrule(lr){6-9}\cmidrule(lr){10-13}\cmidrule(l){14-17}
Method & FID$\downarrow$ & IS$\uparrow$ & Prec.$\uparrow$ & Rec.$\uparrow$ & FID$\downarrow$ & IS$\uparrow$ & Prec.$\uparrow$ & Rec.$\uparrow$ & FID$\downarrow$ & IS$\uparrow$ & Prec.$\uparrow$ & Rec.$\uparrow$ & FID$\downarrow$ & IS$\uparrow$ & Prec.$\uparrow$ & Rec.$\uparrow$\\\midrule
\rowcolor{benchmarkGreen}
SiT & 59.272 & 23.753 & 0.402 & 0.588 & 35.789 & 41.626 & 0.526 & 0.636 & 19.884 & 68.692 & 0.628 & 0.637 & 16.685 & 78.990 & 0.652 & 0.630\\
REPA & 49.648 & 29.799 & 0.448 & \textbf{0.627} & 21.960 & 67.625 & 0.599 & 0.648 & 10.040 & 111.154 & 0.685 & \textbf{0.659} & 7.791 & 126.842 & 0.702 & \textbf{0.659}\\
iREPA & \underline{45.203} & \underline{34.098} & 0.476 & \underline{0.618} & 20.840 & \underline{71.556} & 0.602 & \underline{0.649} & 9.155 & 116.127 & 0.689 & 0.650 & 7.822 & 125.936 & \underline{0.703} & \underline{0.657}\\
HASTE & 45.322 & 33.019 & \underline{0.476} & 0.615 & \underline{20.780} & 69.432 & \textbf{0.614} & 0.638 & \textbf{8.425} & \underline{120.921} & \textbf{0.702} & \underline{0.652} & \underline{7.288} & \underline{128.941} & \textbf{0.710} & 0.651\\
\addlinespace[2pt]\rowcolor{benchmarkBlue}
\textbf{\method} & \textbf{44.109} & \textbf{35.735} & \textbf{0.477} & 0.615 & \textbf{19.848} & \textbf{74.670} & \underline{0.609} & \textbf{0.651} & \underline{8.430} & \textbf{124.335} & \underline{0.694} & 0.651 & \textbf{7.239} & \textbf{133.547} & 0.701 & 0.655\\
\addlinespace[5pt]\midrule
\textbf{CFG = 1.5} & \multicolumn{4}{c}{\textbf{SiT-S/2}} & \multicolumn{4}{c}{\textbf{SiT-B/2}} & \multicolumn{4}{c}{\textbf{SiT-L/2}} & \multicolumn{4}{c}{\textbf{SiT-XL/2}}\\
\cmidrule(lr){2-5}\cmidrule(lr){6-9}\cmidrule(lr){10-13}\cmidrule(l){14-17}
Method & FID$\downarrow$ & IS$\uparrow$ & Prec.$\uparrow$ & Rec.$\uparrow$ & FID$\downarrow$ & IS$\uparrow$ & Prec.$\uparrow$ & Rec.$\uparrow$ & FID$\downarrow$ & IS$\uparrow$ & Prec.$\uparrow$ & Rec.$\uparrow$ & FID$\downarrow$ & IS$\uparrow$ & Prec.$\uparrow$ & Rec.$\uparrow$\\\midrule
\rowcolor{benchmarkGreen}
SiT & 30.809 & 51.463 & 0.562 & \underline{0.543} & 12.501 & 107.426 & 0.725 & 0.520 & 5.290 & 176.550 & 0.821 & 0.496 & 4.360 & 197.761 & 0.839 & 0.501\\
REPA & 23.257 & 69.789 & 0.615 & \textbf{0.548} & 5.347 & 178.461 & 0.784 & 0.530 & \underline{2.855} & 262.805 & 0.847 & 0.533 & \underline{2.906} & 285.662 & \underline{0.856} & \underline{0.534}\\
iREPA & \underline{18.794} & \underline{85.190} & \underline{0.647} & 0.537 & \underline{4.800} & \underline{187.990} & \underline{0.791} & \underline{0.535} & 2.888 & 271.820 & \underline{0.851} & \underline{0.534} & 2.926 & 287.120 & 0.855 & 0.531\\
HASTE & 18.936 & 81.210 & \textbf{0.661} & 0.517 & 5.020 & 181.923 & \textbf{0.808} & 0.505 & 2.944 & \underline{276.692} & \textbf{0.857} & 0.521 & 3.060 & \textbf{292.904} & \textbf{0.860} & 0.529\\
\addlinespace[2pt]\rowcolor{benchmarkOrange}
\textbf{\method} & \textbf{18.524} & \textbf{87.835} & 0.641 & 0.543 & \textbf{4.678} & \textbf{191.507} & 0.787 & \textbf{0.537} & \textbf{2.762} & \textbf{279.825} & 0.847 & \textbf{0.541} & \textbf{2.728} & \underline{292.146} & 0.848 & \textbf{0.549}\\
\bottomrule\end{tabularx}\endgroup
\par\vspace{8pt}
\centering
\includegraphics[width=\textwidth]{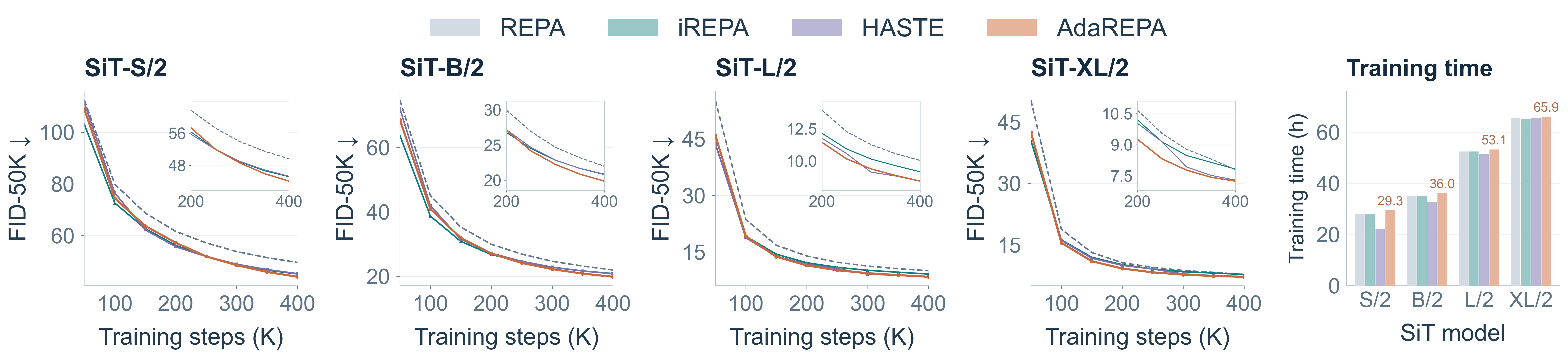}
\captionof{figure}{\textbf{Generation quality and training time across SiT scales.} Left: FID-50K ($\downarrow$) at CFG scale 1, with insets enlarging 200K--400K. Right: Training time in hours for 400K steps.}
\label{fig:main-convergence}
\end{table*}

\subsection{Experimental Setup}
\label{sec:setup}
\paragraph{Models and Datasets.}
Following SiT~\cite{sit} and REPA~\cite{repa}, we use SiT-S/2, B/2, L/2, and XL/2 with patch size 2 on ImageNet~\cite{imagenet} at $256\times256$ resolution. These models have 12, 12, 24, and 28 layers, with hidden dimensions of 384, 768, 1024, and 1152, respectively. As in REPA, we use SD-VAE-ft-MSE to encode training images into $32\times32\times4$ latents and SD-VAE-ft-EMA to decode generated samples. The frozen DINOv2 ViT-B/14 teacher~\cite{dino} produces $16\times16\times768$ target features from $224\times224$ clean images.

\paragraph{Evaluation Metrics.}
For the main comparisons in Table~\ref{tab:benchmark} and Fig.~\ref{fig:main-convergence}, we generate 50K images from EMA model weights. We report FID~\cite{fid}, Inception Score (IS), Precision (Prec.), and Recall (Rec.) in the table, and FID in the training curves. Images are generated with a 250-step Euler--Maruyama SDE sampler, using CFG scales 1 and 1.5 for the table and scale 1 for the curves. FID is computed against the ADM reference statistics~\cite{adm}, with the sample count fixed within each comparison.

\paragraph{Baselines.}
We compare \method with four baselines, all using the same SiT backbone at each model scale. SiT~\cite{sit} trains the generative backbone without teacher supervision. REPA~\cite{repa} aligns a fixed student layer with features from a frozen visual teacher. iREPA~\cite{irepa} replaces the MLP projector with a convolutional projector and spatially normalizes teacher features to better preserve spatial structure. HASTE~\cite{haste} adds attention alignment to REPA and stops both alignment losses partway through training.

\paragraph{Implementation Details.}
For the main comparison, we train all four SiT scales for 400K steps with a global batch size of 256, corresponding to 102.4M training images. During training, class labels are dropped with probability 0.1, and we maintain an exponential moving average (EMA) of the model parameters with decay 0.9999. We follow the configurations reported in the respective papers unless otherwise specified. REPA uses an MLP projector and an alignment coefficient of 0.5, with supervision at Layer 4 for S/2 and B/2, and Layer 8 for L/2 and XL/2. For HASTE, the feature and attention alignment coefficients are both 0.5; both losses are terminated at 100K steps for S/2 and B/2, and 250K for L/2 and XL/2. iREPA uses convolutional projection and spatial normalization with an alignment coefficient of 1.0. \method uses a separate projector for each layer, with $\lambda=0.5$ and a score EMA coefficient of $\beta=0.99$. Projectors are discarded after training and add no inference cost. These experiments use linear interpolants with $t\sim\mathcal{U}(0,1)$, velocity prediction, and AdamW (learning rate $10^{-4}$, optimizer coefficients $(\beta_1,\beta_2)=(0.9,0.999)$, zero weight decay), with FP16 and gradient clipping at 1.0.

\subsection{Main Results}
\label{sec:quality}
Table~\ref{tab:benchmark} compares generation quality across four SiT scales at 400K training steps, with and without classifier-free guidance. At CFG scale 1, \method achieves the lowest FID on S/2, B/2, and XL/2, and the highest IS at all four scales. On L/2, its FID of 8.430 is close to HASTE's 8.425. At CFG scale 1.5, \method achieves the lowest FID across all four scales, reaching 18.524, 4.678, 2.762, and 2.728 on S/2, B/2, L/2, and XL/2, respectively. It also gives the highest IS on S/2, B/2, and L/2, and the highest recall on B/2, L/2, and XL/2. The improvements over REPA across both guidance settings show that adapting the supervision layer benefits students of different capacities.

Figure~\ref{fig:main-convergence} compares the training curves across all four SiT scales. \method gives lower FID than REPA throughout the plotted training period. The insets show its advantage over the other alignment methods at 400K on S/2, B/2, and XL/2, while its final FID on L/2 is close to HASTE.

\method concentrates adaptation on the allocation of supervision across layers, activating one alignment branch per step and keeping the same single-layer supervision budget as REPA. Under this budget, it improves generation quality with only a minimal increase in training time due to gradient scoring, which becomes increasingly negligible at larger model scales (Fig.~\ref{fig:main-convergence}, right). The relative increase is particularly small on XL/2. All projectors are removed after training, so generation uses the original SiT backbone and sampler with no additional inference cost.

\subsection{Supervision Dynamics over Training}
\label{sec:routing-dynamics}
\begin{figure}[t]
\centering
\includegraphics[width=\linewidth]{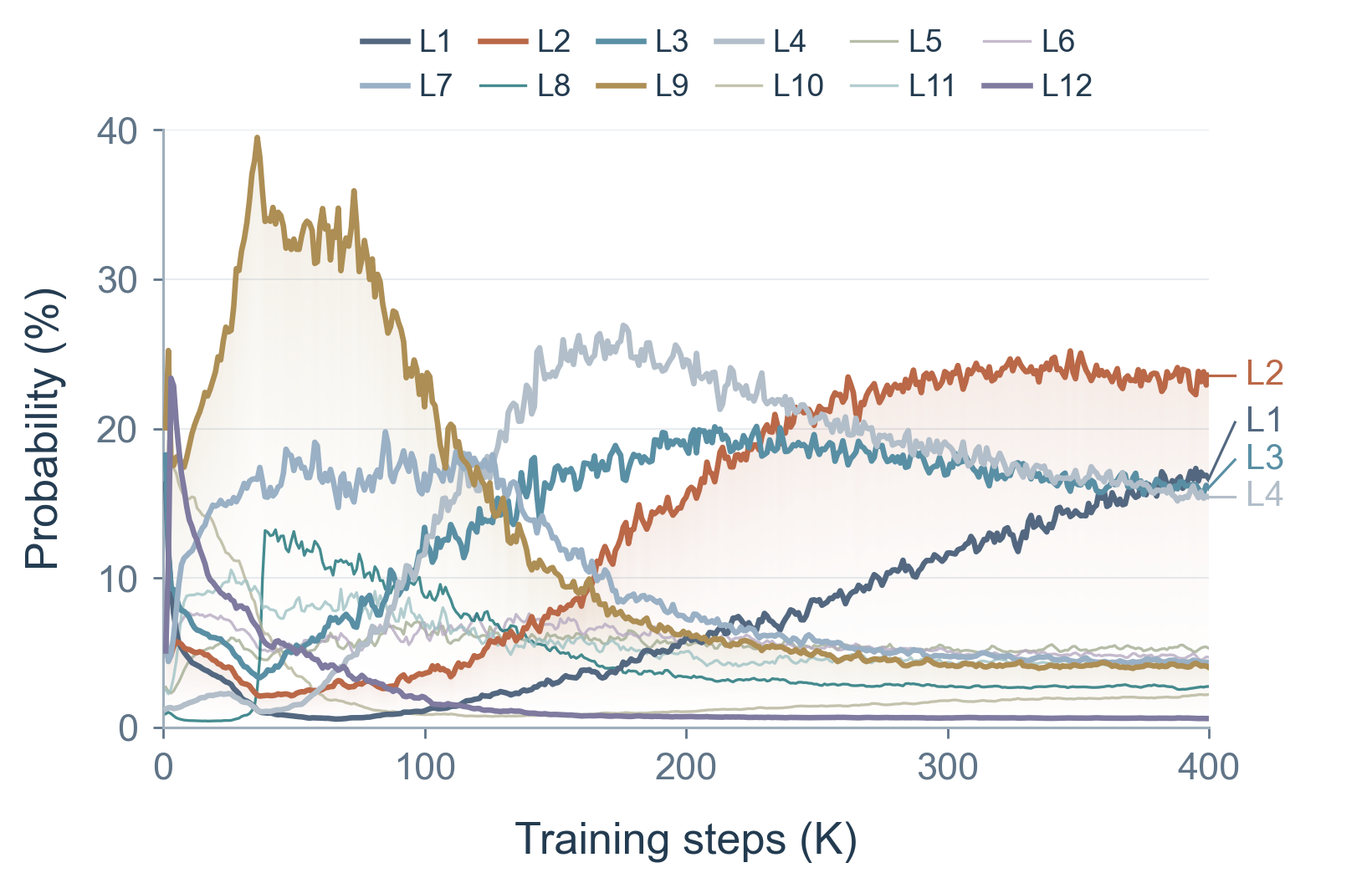}
\setlength{\abovecaptionskip}{3pt}
\caption{\textbf{Routing probabilities over training on SiT-S/2.} The dominant position shifts from Layer 9 to Layer 4 and then Layer 2, with overlapping contributions from other layers.}
\label{fig:layer-probabilities}
\end{figure}

Figure~\ref{fig:layer-probabilities} shows how supervision preferences change over training. Layer 9 receives the highest probability early on, peaking near $40\%$ before declining. Layer 4 becomes dominant in the middle of training, while Layer 2 gradually gains probability and leads later. These changes reveal an overall deep-to-shallow trend without a preset switching schedule.

The changes overlap rather than form separate phases. As Layer 9 declines, Layers 3 and 4 gain probability. Layer 2 then rises while Layer 4 remains active, and Layer 1 continues to gain probability late in training. By 400K, Layer 2 leads, while Layers 1, 3, and 4 retain substantial probabilities. The later distribution therefore favors several relatively shallow positions rather than settling on one fixed layer.

The early preference for Layer 9 may reflect a greater need to establish semantic representations, consistent with the hypothesis in Sec.~\ref{sec:motivation}. As these representations develop, moving supervision toward shallow layers leaves more subsequent blocks for generative refinement. Layer 4's temporary dominance suggests that an intermediate position can offer a useful compromise during this transition, while the favored position continues to change.

\subsection{Ablation Studies}
\label{sec:design}
We first compare teacher models, then examine routing granularity and design choices.

\subsubsection{Different Teacher Models}
\label{sec:teacher-ablation}
We compare REPA and \method with three teachers: DINOv2 ViT-B/14~\cite{dino}, CLIP ViT-L/14~\cite{clip}, and DINOv3 ViT-B/16~\cite{dinov3}. All experiments use SiT-B/2, with the remaining training and evaluation settings following the main comparison in Sec.~\ref{sec:setup}. We report FID on 50K samples from EMA weights at CFG scale 1.

\begin{figure}[t]
\centering
\includegraphics[width=\linewidth]{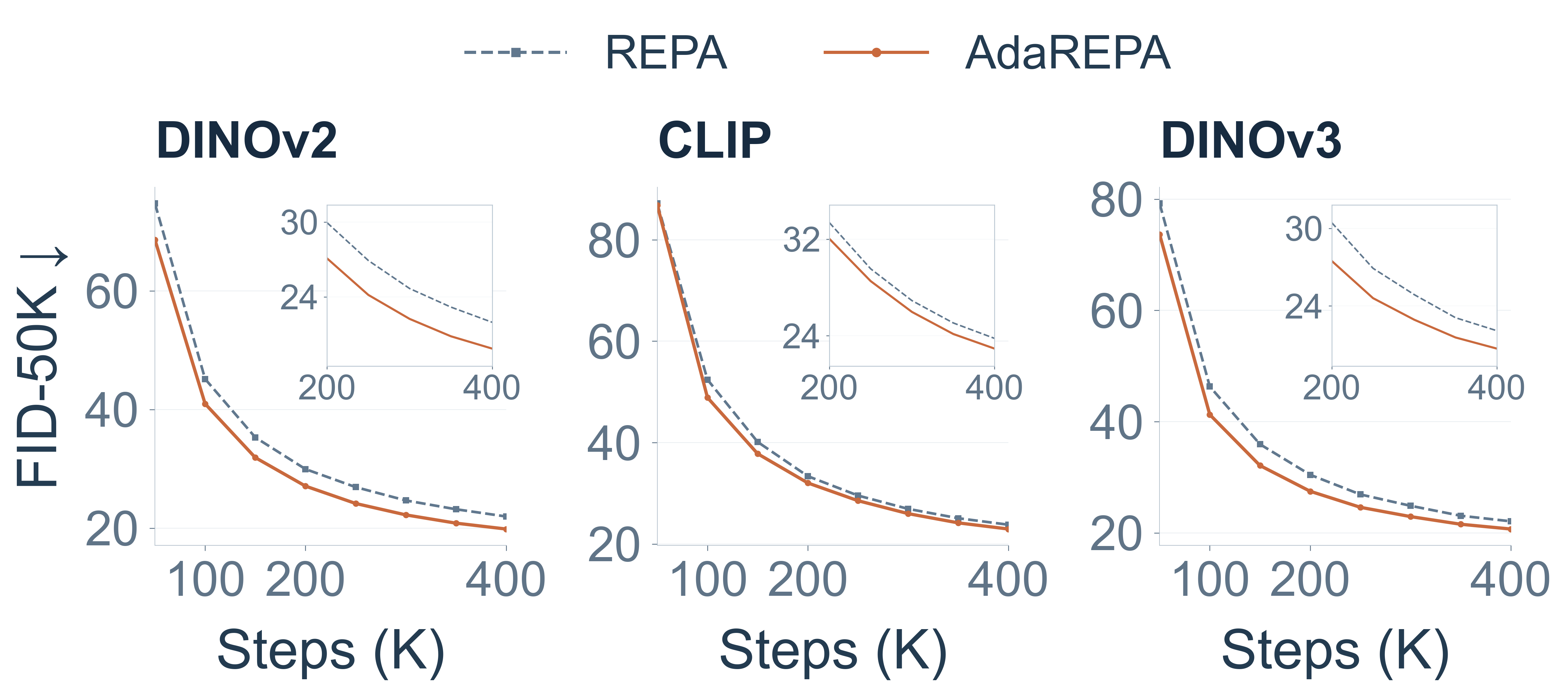}
\caption{\textbf{Different teacher models.} FID ($\downarrow$) on SiT-B/2 with three teachers at CFG scale 1. Insets enlarge the 200K--400K interval.}
\label{fig:teacher-comparison}
\end{figure}

Figure~\ref{fig:teacher-comparison} shows that \method achieves lower FID than fixed-layer REPA with all three teachers throughout the plotted training period. The advantage remains at 400K, although the size of the gap varies across teachers. These results extend the benefit of adaptive layer selection beyond the DINOv2 teacher used in the main comparison: as the teacher representation changes, adapting where its guidance enters the student continues to improve generation.

\begin{figure}[t]
\centering
\includegraphics[width=\linewidth]{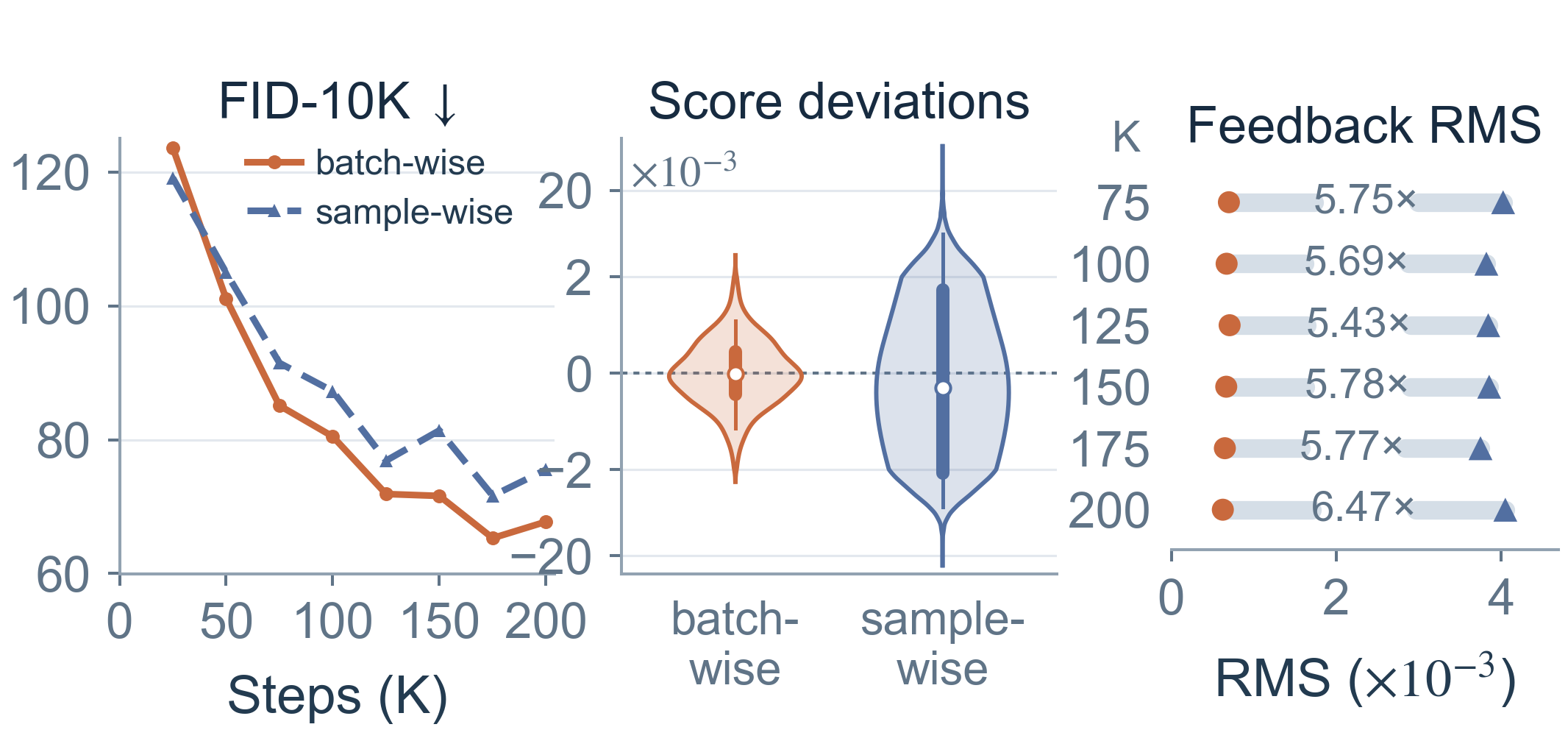}
\setlength{\abovecaptionskip}{3pt}
\caption{\textbf{Effect of selection granularity.} Left: FID-10K. Middle: Score deviations from the pre-update EMA, with medians and interquartile ranges on a symmetric log scale. Right: Deviation RMS over 25K intervals; labels show interval endpoints and sample-wise/batch-wise ratios.}
\label{fig:routing-granularity}
\end{figure}

\begin{figure*}[t]
\centering\includegraphics[width=\textwidth]{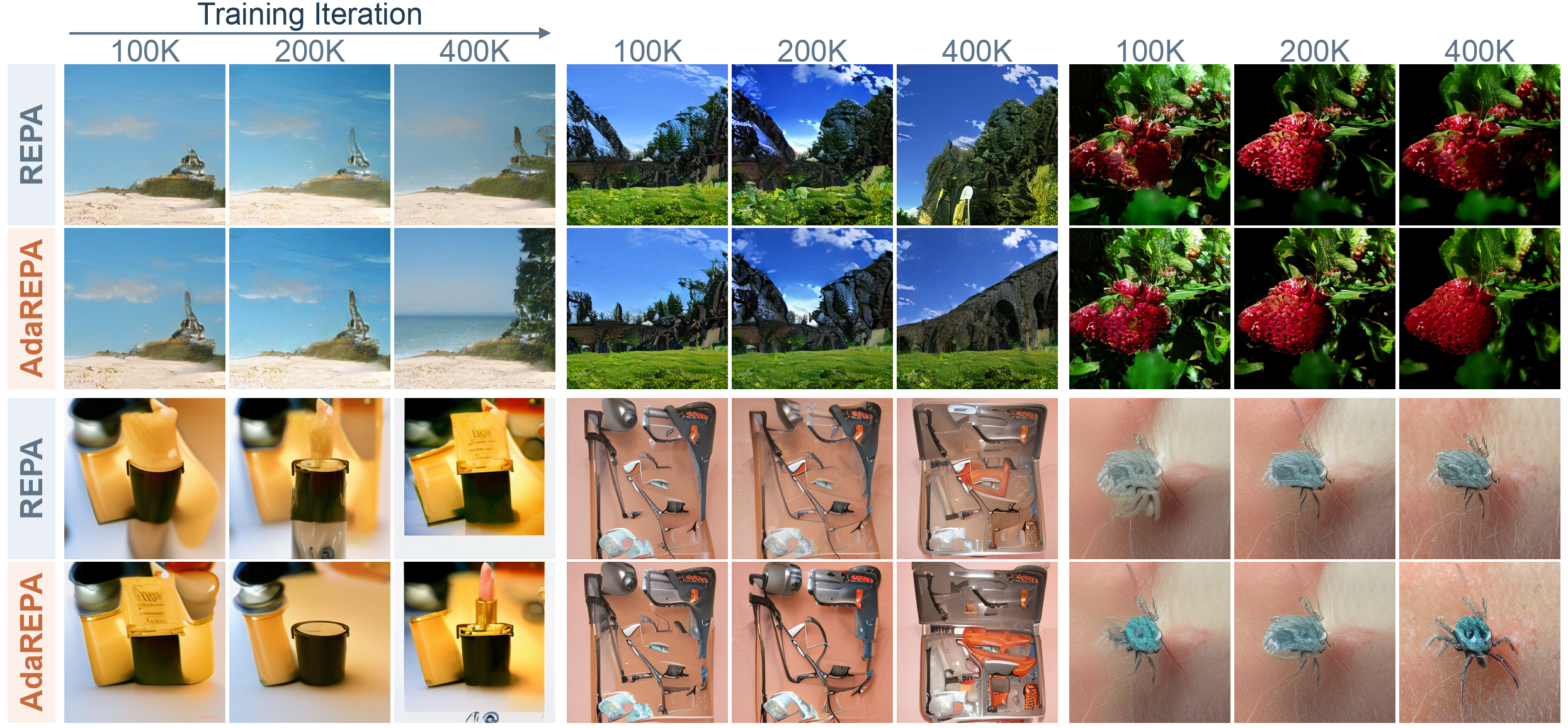}
\caption{\textbf{ImageNet-256 samples over training.} Rows 1--2 show SiT-S/2, while rows 3--4 show SiT-XL/2, all at CFG scale 1. Within each pair, REPA is followed by \method. Columns within each group show 100K, 200K, and 400K training steps.}
\label{fig:samples}
\end{figure*}

\subsubsection{Batch-wise vs. Sample-wise Selection}
\label{sec:sampling-ablation}
We compare batch-wise and sample-wise layer selection on SiT-S/2. At the same global batch size, sample-wise selection supervises more layers within a step but splits the images used to score each layer into smaller subsets. Batch-wise selection uses the full batch to form feedback at one layer and explores different layers through sampling across steps. It achieves lower FID at every evaluated checkpoint from 50K to 200K (Fig.~\ref{fig:routing-granularity}, left).

To examine feedback stability, we measure the difference between each observed gradient score and its smoothed score before the update. Sample-wise selection produces a wider deviation distribution (Fig.~\ref{fig:routing-granularity}, middle), with a deviation RMS $5.43$--$6.47$ times that of batch-wise selection across six consecutive 25K intervals (right). Even for the same layer at a fixed model state, different image subsets can give different scores because gradient agreement varies across images. With fewer images per layer, replacing a few images can shift the score and alter which layer appears more suitable. This can lead to larger fluctuations in layer scores, which then affect supervision for subsequent batches through the EMA and sampling probabilities. Batch-wise selection uses the full batch to form each score, helping reduce this sensitivity while retaining exploration across steps. More stable feedback helps the sampling probabilities reflect layer suitability at the current training stage and adjust supervision as the student's learning needs change. Both results are consistent with this explanation.

\begin{table}[t]
\caption{\textbf{Gradient metrics and projector design.} SiT-S/2 FID-10K ($\downarrow$). Each block varies one component.}
\label{tab:abl}\centering
\small
\begin{tabular*}{\columnwidth}{@{\extracolsep{\fill}}lrrrr@{}}\toprule
Variant & 100K & 200K & 300K & 400K\\\midrule
\multicolumn{5}{@{}l}{\textit{(a) Gradient metric: independent projectors}}\\
Mean Cosine & \textbf{79.680} & 68.320 & 56.179 & 47.562\\
BP-Cosine & 79.990 & 68.942 & 56.163 & 48.146\\
CS-Cosine & 80.473 & \textbf{67.747} & \textbf{56.020} & \textbf{46.940}\\\midrule
\multicolumn{5}{@{}l}{\textit{(b) Projectors}}\\
Shared & 85.968 & 71.695 & 58.686 & 51.119\\
Independent & \textbf{80.473} & \textbf{67.747} & \textbf{56.020} & \textbf{46.940}\\\bottomrule
\end{tabular*}
\end{table}

\subsubsection{Design Choices}
\label{sec:score-analysis}
\paragraph{Gradient aggregation.}
The aggregation rule determines how samples contribute to the layer score. With independent projectors, Mean Cosine gives the lowest FID at 100K, while CS-Cosine performs best at 200K, 300K, and 400K. At 400K, CS-Cosine reaches 46.940, compared with 47.562 for Mean Cosine and 48.146 for BP-Cosine (Table~\ref{tab:abl}).
\par
\begingroup\widowpenalty=10000
\noindent\textbf{Independent projectors.}
Independent projectors reduce FID at all four checkpoints, reaching 46.940 at 400K compared with 51.119 for a shared projector (Table~\ref{tab:abl}). With a shared projector, an update from one layer also changes the mapping used by others. Separate projectors learn each layer's mapping to the teacher as routing changes.
\par\endgroup

\subsection{Qualitative Results}
\label{sec:qualitative}
Figure~\ref{fig:samples} compares REPA and \method over training on SiT-S/2 and SiT-XL/2 at CFG scale 1. Across both model sizes, \method improves how local parts and textures are organized into coherent scenes and objects. At 400K, the viaduct has more continuous arches, and tools in the case have clearer handles and boundaries. In the seashore example, the anomalous central structure disappears, revealing a clearer boundary between sea and land. The lipstick forms a recognizable tip seated in its holder, while the arthropod develops a clearer body outline and more distinct legs. The strawberry also gains a more complete shape while retaining its surface texture. Over training, local details become better integrated into coherent scenes and objects.

This joint improvement in structure and appearance is consistent with the routing dynamics in Sec.~\ref{sec:routing-dynamics}. As semantic representations become more established, supervision shifts toward shallower layers, introducing teacher guidance earlier in the network and leaving more subsequent blocks for transformations driven by the generative objective. This may help the student use semantic guidance to maintain subject structure while the generative objective further refines local shapes and textures. The examples at both model sizes reflect this balance between semantic learning and generative refinement, with gains in both structure and local detail.

\section{Conclusion}
\label{sec:conclusion}
In this work, we investigated where teacher supervision should enter a Diffusion Transformer as its internal representations evolve. We found that the relative benefits of different supervision depths change across training stages, challenging the conventional practice of fixed-layer alignment. Motivated by this finding, we introduced \method, an online routing framework that adapts the supervision depth using gradient compatibility between the flow-matching and alignment objectives. Experiments on ImageNet-256 across four SiT model sizes and multiple visual teachers demonstrate improved convergence and generation quality over fixed-layer REPA, with minimal additional training overhead and no inference cost. The learned routing adapts supervision preferences throughout training without relying on a predefined schedule. These findings suggest that the supervision layer should adapt to the student's training state rather than remain fixed throughout training.

{\small\bibliographystyle{ieeenat_fullname}\bibliography{bib/references}}

\begin{thebibliography}{27}
\providecommand{\natexlab}[1]{#1}
\providecommand{\url}[1]{\texttt{#1}}
\expandafter\ifx\csname urlstyle\endcsname\relax
  \providecommand{\doi}[1]{doi: #1}\else
  \providecommand{\doi}{doi: \begingroup \urlstyle{rm}\Url}\fi

\bibitem[Chen et~al.(2021{\natexlab{a}})Chen, Mei, Zhang, Wang, Wang, Feng, and
  Chen]{semckd}
Defang Chen, Jian-Ping Mei, Yuan Zhang, Can Wang, Zhe Wang, Yan Feng, and Chun
  Chen.
\newblock Cross-layer distillation with semantic calibration.
\newblock In \emph{AAAI Conference on Artificial Intelligence},
  2021{\natexlab{a}}.

\bibitem[Chen et~al.(2021{\natexlab{b}})Chen, Liu, Zhao, and Jia]{reviewkd}
Pengguang Chen, Shu Liu, Hengshuang Zhao, and Jiaya Jia.
\newblock Distilling knowledge via knowledge review.
\newblock In \emph{IEEE/CVF Conference on Computer Vision and Pattern
  Recognition}, pages 5008--5017, 2021{\natexlab{b}}.

\bibitem[Chen et~al.(2018)Chen, Badrinarayanan, Lee, and Rabinovich]{gradnorm}
Zhao Chen, Vijay Badrinarayanan, Chen-Yu Lee, and Andrew Rabinovich.
\newblock {GradNorm}: Gradient normalization for adaptive loss balancing in
  deep multitask networks.
\newblock In \emph{International Conference on Machine Learning}, pages
  794--803, 2018.

\bibitem[Deng et~al.(2009)Deng, Dong, Socher, Li, Li, and Fei-Fei]{imagenet}
Jia Deng, Wei Dong, Richard Socher, Li-Jia Li, Kai Li, and Li Fei-Fei.
\newblock {ImageNet}: A large-scale hierarchical image database.
\newblock In \emph{IEEE Conference on Computer Vision and Pattern Recognition},
  2009.

\bibitem[Dhariwal and Nichol(2021)]{adm}
Prafulla Dhariwal and Alexander Nichol.
\newblock Diffusion models beat {GANs} on image synthesis.
\newblock In \emph{Advances in Neural Information Processing Systems}, 2021.

\bibitem[Du et~al.(2018)Du, Czarnecki, Jayakumar, Farajtabar, Pascanu, and
  Lakshminarayanan]{du}
Yunshu Du, Wojciech~M. Czarnecki, Siddhant~M. Jayakumar, Mehrdad Farajtabar,
  Razvan Pascanu, and Balaji Lakshminarayanan.
\newblock Adapting auxiliary losses using gradient similarity.
\newblock \emph{arXiv preprint arXiv:1812.02224}, 2018.

\bibitem[Heusel et~al.(2017)Heusel, Ramsauer, Unterthiner, Nessler, and
  Hochreiter]{fid}
Martin Heusel, Hubert Ramsauer, Thomas Unterthiner, Bernhard Nessler, and Sepp
  Hochreiter.
\newblock {GANs} trained by a two time-scale update rule converge to a local
  nash equilibrium.
\newblock In \emph{Advances in Neural Information Processing Systems}, 2017.

\bibitem[Lee et~al.(2015)Lee, Xie, Gallagher, Zhang, and Tu]{dsn}
Chen-Yu Lee, Saining Xie, Patrick Gallagher, Zhengyou Zhang, and Zhuowen Tu.
\newblock Deeply-supervised nets.
\newblock In \emph{International Conference on Artificial Intelligence and
  Statistics}, pages 562--570, 2015.

\bibitem[Lipman et~al.(2023)Lipman, Chen, Ben-Hamu, Nickel, and Le]{fm}
Yaron Lipman, Ricky T.~Q. Chen, Heli Ben-Hamu, Maximilian Nickel, and Matt Le.
\newblock Flow matching for generative modeling.
\newblock In \emph{International Conference on Learning Representations}, 2023.

\bibitem[Liu et~al.(2021)Liu, Liu, Jin, Stone, and Liu]{cagrad}
Bo Liu, Xingchao Liu, Xiaojie Jin, Peter Stone, and Qiang Liu.
\newblock Conflict-averse gradient descent for multi-task learning.
\newblock In \emph{Advances in Neural Information Processing Systems}, 2021.

\bibitem[Liu et~al.(2023)Liu, Gong, and Liu]{rectified}
Xingchao Liu, Chengyue Gong, and Qiang Liu.
\newblock Flow straight and fast: Learning to generate and transfer data with
  rectified flow.
\newblock In \emph{International Conference on Learning Representations}, 2023.

\bibitem[Ma et~al.(2024)Ma, Goldstein, Albergo, Boffi, Vanden-Eijnden, and
  Xie]{sit}
Nanye Ma, Mark Goldstein, Michael~S. Albergo, Nicholas~M. Boffi, Eric
  Vanden-Eijnden, and Saining Xie.
\newblock {SiT}: Exploring flow and diffusion-based generative models with
  scalable interpolant transformers.
\newblock In \emph{European Conference on Computer Vision}, 2024.

\bibitem[Oquab et~al.(2024)Oquab, Darcet, Moutakanni, Vo, Szafraniec, Khalidov,
  Fernandez, Haziza, Massa, El-Nouby, Assran, Ballas, Galuba, Howes, Huang, Li,
  Misra, Rabbat, Sharma, Synnaeve, Xu, J{\'e}gou, Mairal, Labatut, Joulin, and
  Bojanowski]{dino}
Maxime Oquab, Timoth{\'e}e Darcet, Th{\'e}o Moutakanni, Huy~V. Vo, Marc
  Szafraniec, Vasil Khalidov, Pierre Fernandez, Daniel Haziza, Francisco Massa,
  Alaaeldin El-Nouby, Mahmoud Assran, Nicolas Ballas, Wojciech Galuba, Russell
  Howes, Po-Yao Huang, Shang-Wen Li, Ishan Misra, Michael Rabbat, Vasu Sharma,
  Gabriel Synnaeve, Hu Xu, Herv{\'e} J{\'e}gou, Julien Mairal, Patrick Labatut,
  Armand Joulin, and Piotr Bojanowski.
\newblock {DINOv2}: Learning robust visual features without supervision.
\newblock \emph{Transactions on Machine Learning Research}, 2024.

\bibitem[Passban et~al.(2021)Passban, Wu, Rezagholizadeh, and Liu]{alpkd}
Peyman Passban, Yimeng Wu, Mehdi Rezagholizadeh, and Qun Liu.
\newblock {ALP-KD}: Attention-based layer projection for knowledge
  distillation.
\newblock In \emph{AAAI Conference on Artificial Intelligence}, 2021.

\bibitem[Peebles and Xie(2023)]{dit}
William Peebles and Saining Xie.
\newblock Scalable diffusion models with transformers.
\newblock In \emph{IEEE/CVF International Conference on Computer Vision}, pages
  4195--4205, 2023.

\bibitem[Radford et~al.(2021)Radford, Kim, Hallacy, Ramesh, Goh, Agarwal,
  Sastry, Askell, Mishkin, Clark, Krueger, and Sutskever]{clip}
Alec Radford, Jong~Wook Kim, Chris Hallacy, Aditya Ramesh, Gabriel Goh,
  Sandhini Agarwal, Girish Sastry, Amanda Askell, Pamela Mishkin, Jack Clark,
  Gretchen Krueger, and Ilya Sutskever.
\newblock Learning transferable visual models from natural language
  supervision.
\newblock In \emph{International Conference on Machine Learning}, pages
  8748--8763, 2021.

\bibitem[Romero et~al.(2015)Romero, Ballas, Ebrahimi~Kahou, Chassang, Gatta,
  and Bengio]{fitnets}
Adriana Romero, Nicolas Ballas, Samira Ebrahimi~Kahou, Antoine Chassang, Carlo
  Gatta, and Yoshua Bengio.
\newblock {FitNets}: Hints for thin deep nets.
\newblock In \emph{International Conference on Learning Representations}, 2015.

\bibitem[Sener and Koltun(2018)]{mgda}
Ozan Sener and Vladlen Koltun.
\newblock Multi-task learning as multi-objective optimization.
\newblock In \emph{Advances in Neural Information Processing Systems}, 2018.

\bibitem[Sim{\'e}oni et~al.(2025)Sim{\'e}oni, Vo, Seitzer, Baldassarre, Oquab,
  Jose, Khalidov, Szafraniec, Yi, Ramamonjisoa, Massa, Haziza, Wehrstedt, Wang,
  Darcet, Moutakanni, Sentana, Roberts, Vedaldi, Tolan, Brandt, Couprie,
  Mairal, J{\'e}gou, Labatut, and Bojanowski]{dinov3}
Oriane Sim{\'e}oni, Huy~V. Vo, Maximilian Seitzer, Federico Baldassarre, Maxime
  Oquab, Cijo Jose, Vasil Khalidov, Marc Szafraniec, Seungeun Yi, Micha{\"e}l
  Ramamonjisoa, Francisco Massa, Daniel Haziza, Luca Wehrstedt, Jianyuan Wang,
  Timoth{\'e}e Darcet, Th{\'e}o Moutakanni, Leonel Sentana, Claire Roberts,
  Andrea Vedaldi, Jamie Tolan, John Brandt, Camille Couprie, Julien Mairal,
  Herv{\'e} J{\'e}gou, Patrick Labatut, and Piotr Bojanowski.
\newblock {DINOv3}.
\newblock \emph{arXiv preprint arXiv:2508.10104}, 2025.

\bibitem[Singh et~al.(2026)Singh, Leng, Wu, Zheng, Zhang, Shechtman, and
  Xie]{irepa}
Jaskirat Singh, Xingjian Leng, Zongze Wu, Liang Zheng, Richard Zhang, Eli
  Shechtman, and Saining Xie.
\newblock What matters for representation alignment: Global information or
  spatial structure?
\newblock In \emph{International Conference on Learning Representations}, 2026.

\bibitem[Wang et~al.(2026)Wang, Chen, Fan, Jiang, Zhang, Sun, Wang, Lu, and
  Shen]{rare}
Yongcong Wang, Hingchin Chen, Mingyu Fan, Shuo Jiang, Teer Zhang, Yucong Sun,
  Zijia Wang, Yiming Lu, and Chengchao Shen.
\newblock What visual generators need from teachers: Rethinking representation
  alignment.
\newblock \emph{arXiv preprint arXiv:2609.34732}, 2026.

\bibitem[Wang et~al.(2021)Wang, Tsvetkov, Firat, and Cao]{gradvaccine}
Zirui Wang, Yulia Tsvetkov, Orhan Firat, and Yuan Cao.
\newblock Gradient vaccine: Investigating and improving multi-task optimization
  in massively multilingual models.
\newblock In \emph{International Conference on Learning Representations}, 2021.

\bibitem[Wang et~al.(2025)Wang, Zhao, Zhou, Li, Liang, Shi, Zhao, Zhou, Zhang,
  Wang, Wang, and You]{haste}
Ziqiao Wang, Wangbo Zhao, Yuhao Zhou, Zekai Li, Zhiyuan Liang, Mingjia Shi,
  Xuanlei Zhao, Pengfei Zhou, Kaipeng Zhang, Zhangyang Wang, Kai Wang, and Yang
  You.
\newblock {REPA} works until it doesn't: Early-stopped, holistic alignment
  supercharges diffusion training.
\newblock In \emph{Advances in Neural Information Processing Systems}, 2025.

\bibitem[Yu et~al.(2025)Yu, Kwak, Jang, Jeong, Huang, Shin, and Xie]{repa}
Sihyun Yu, Sangkyung Kwak, Huiwon Jang, Jongheon Jeong, Jonathan Huang, Jinwoo
  Shin, and Saining Xie.
\newblock Representation alignment for generation: Training diffusion
  transformers is easier than you think.
\newblock In \emph{International Conference on Learning Representations}, 2025.

\bibitem[Yu et~al.(2020)Yu, Kumar, Gupta, Levine, Hausman, and Finn]{pcgrad}
Tianhe Yu, Saurabh Kumar, Abhishek Gupta, Sergey Levine, Karol Hausman, and
  Chelsea Finn.
\newblock Gradient surgery for multi-task learning.
\newblock In \emph{Advances in Neural Information Processing Systems}, pages
  5824--5836, 2020.

\bibitem[Zagoruyko and Komodakis(2017)]{attentiontransfer}
Sergey Zagoruyko and Nikos Komodakis.
\newblock Paying more attention to attention: Improving the performance of
  convolutional neural networks via attention transfer.
\newblock In \emph{International Conference on Learning Representations}, 2017.

\bibitem[Zhang et~al.(2026)Zhang, Xie, Yang, and Liu]{agrepa}
Pengfei Zhang, Tianxin Xie, Minghao Yang, and Li Liu.
\newblock {AG-REPA}: Causal layer selection for representation alignment in
  audio flow matching.
\newblock In \emph{International Conference on Machine Learning}, pages
  159454--159470, 2026.

\end{thebibliography}
\end{document}